\documentclass[aps,prb,twocolumn,superscriptaddress,floatfix,longbibliography,nofootinbib]{revtex4-2}%

\usepackage{amsmath,amssymb} 

\DeclareMathOperator*{\argmax}{arg\,max}
\DeclareMathOperator*{\argmin}{arg\,min}
\usepackage[english]{babel}
\usepackage{blindtext}
\usepackage{bm} 
\usepackage{graphicx} 
\usepackage{comment} 
\usepackage{textcomp} 

\usepackage{enumitem}
\usepackage[ruled, lined, linesnumbered, commentsnumbered, longend]{algorithm2e}
\setlist{noitemsep,leftmargin=*,topsep=0pt,parsep=0pt}
\usepackage{lipsum, babel}
\usepackage{float}
\usepackage{svg}
\usepackage{notoccite}
\usepackage[normalem]{ulem}
\usepackage{xcolor} 
\definecolor{lightgray}{gray}{0.6}
\definecolor{medgray}{gray}{0.4}
\usepackage[bottom]{footmisc}
\usepackage{hyperref}
\hypersetup{
colorlinks=true,
urlcolor= blue,
citecolor=blue,
linkcolor= blue,
}

\newcommand\blfootnote[1]{%
  \begingroup
  \renewcommand\thefootnote{}\footnote{#1}%
  \addtocounter{footnote}{-1}%
  \endgroup
}
\newif\ifptitle
\newif\ifpnumber
\newcounter{para}

\ptitletrue  
\pnumbertrue  

\newcommand{\mytitle}{Solving Inverse Problems with Conditional-GAN Prior via Fast Network-Projected Gradient Descent}

\begin{document}

\title{\mytitle}

\newcommand*{\affaddr}[1]{#1}
\newcommand*{\affmark}[1][*]{\textsuperscript{#1}}
\author{
Muhammad Fadli Damara\affmark[1]\affmark[*], Gregor Kornhardt\affmark[1]\affmark[*] and Peter Jung\affmark[1,2]\\
\affaddr{\affmark[1]Technische Universität Berlin, Germany} \\
\affaddr{\affmark[2]Technische Universität München, Germany}\\
}




\blfootnote{*Equal Contribution}\\

\begin{abstract}
The projected gradient descent (PGD) method has shown to be effective
in recovering compressed signals described in a data-driven way by a
generative model, i.e., a generator which has learned the data
distribution. Further reconstruction improvements for such inverse
problems can be achieved by conditioning the generator on the
measurement. The boundary equilibrium generative adversarial network
(BEGAN) implements an equilibrium based loss function and an
auto-encoding discriminator to better balance the performance of the
generator and the discriminator.  In this work we investigate a
network-based projected gradient descent (NPGD) algorithm for measurement-conditional
generative models to solve the inverse problem much faster than
regular PGD. We combine the NPGD with conditional GAN/BEGAN to evaluate
their effectiveness in solving compressed sensing type problems. Our
experiments on the MNIST and CelebA datasets show that the combination
of measurement conditional model with NPGD works well in recovering
the compressed signal while achieving similar or in some cases even
better performance along with a much faster reconstruction.
The achieved reconstruction speed-up in
our experiments is up to $140-175$. 
\end{abstract}

\maketitle

\section{Introduction}\label{sec:Start}
To solve linear inverse problems from compressed sensing measurements, Generative Adversarial Networks (GANs) can be used to describe sophisticated signal structures beyond sparsity in a data driven way. 
Linear inverse problems can be represented using the following linear equations 
\begin{equation}\label{eq:INV}
    y = A x^* + \eta, 
\end{equation}
where $x^{*}\in\mathbb{R}^n$ is a structured target vector to be recovered, $A\in\mathbb{R}^{m\times n}$ is the measurement matrix, and $\eta\in\mathbb{R}^m$ is additive noise. For $m\ll n$ this includes a linear compression directly during the observation. The compressed observation is denoted by $y\in\mathbb{R}^m$.

Due to the restriction $m \ll n$ inverse problem are ill-posed, i.e. the columns of $A$ are linear dependent. For ill-posed problems there are usually multiple solutions, i.e. vectors $x$ which are consistent with the observations in (\ref{eq:INV}). The unknown target vector (ground truth) $x^{*}$ must thus be estimated by solving the following constrained optimization problem: 
\begin{equation}\label{eq:loss_equation}
    \hat{x} = \argmin_{x \in \mathcal{S} }\mathcal{L}(y, Ax),
\end{equation}
where $\mathcal{L}: \mathbb{R}^m \rightarrow \mathbb{R}^+$ is a loss function which measures consistency of $x$ to the measurements $y$, for example a norm $\ell_p$ of the residual $\|y-Ax\|_p$ \cite{shah2018solving}. 
A common way is to use a regularizer which promotes meaningful solutions. However, in this work we focus on an approach where the restriction of $\mathcal{S}\subseteq \mathbb{R}^n$ is used to ensure that it is possible to recover $x^{*}$. In classical compressed sensing, the signal $x^{*}$ has to be (a) sufficiently sparse and (b) the measurement matrix $A$ has to meet conditions such as Restricted Isometry Property (RIP), the nullspace property, or the related Restricted Eigenvalue Condition (REC). For a general overview we refer to the book of Foucart\cite{foucart2013mathematical} and references therein. For example, RIP states that the $\ell_2$-norm of sufficiently sparse vectors is almost preserved under $A$, and thus, given the measurements, robust recovery of sparse signals is guaranteed. More precisely, the isometry constant $\delta_s \in [0,1]$ is the smallest number such that

\begin{align}
    (1- \delta_s) \|x\|^2_2 \leq \|Ax\|^2_2 \leq (1 + \delta_s) \|x\|^2_2 
    \label{eq:rip}
\end{align}

holds for all $x\in\mathcal{S}$, where $\mathcal{S}$ is the set of all $s$-sparse vectors. If $\delta_{2s}<1$ this ensures that a unique $s$-sparse signal exists in the noiseless case. And, $\delta_{2s}<1/\sqrt{2}$ ensures that $s$-sparse $x$ can be recovered by non-NP-hard methods, e.g. convex programs. \cite{Cai2014} This also extends to the noisy setup.

Linear compression during observation has the advantage to substantially reduce complexity, storage costs and communication overhead at the compression stage compared to classical compression algorithms, however the decompression is much more complex, but also comes with many other advantages, important in certain applications. Since (\ref{eq:INV}) is a forward model describing data acquisition, it can be understood as being implemented in the observation device and possibly on the analog level. This methodology is already well-investigated in the last years for many structured data models. In this work, we contribute to the recent developments and study this for structures given by generative models. We investigated the interplay between network architecture, training procedure and decompression algorithms. To examine these interrelationships, we introduce the basic conceptional description of our approach, discuss the algorithms, and highlight also few theoretical results in this context. In the experimental section, we compare the two reconstruction algorithms with unconditional and conditional models and assess their properties.


\section{Generative Adversarial Network}
GANs have proven to be a success for solving inverse problems in a data-driven way due to its powerful generating capabilities. It implicitly learns the data distribution $p_{x}$ (the prior in the inverse problem) from samples, while then creating a computational model to draw samples from. GANs are composed of two models, the generative model $G: \mathbb{R}^k \rightarrow \mathbb{R}^n$ and the discriminative model $D: \mathbb{R}^n \rightarrow \mathbb{R}$. The generator maps a random vector $z \in \mathbb{R}^k$ from a fixed distribution (usually Gaussian distributions $\mathcal{N}(0, I_k)$) to an induced structured data distribution, whereas the discriminator evaluates whether a sample $x$ belongs to the data distribution or the distribution of generated data. These models are usually represented by a neural network. Optimizing both models is equivalent to finding a Nash equilibrium in a min-max game, i.e. a maximum loss is minimized. The competition between the generative model and the discriminative model allows both models to improve \cite{goodfellow2014generative}.

A conditional GAN extends the GAN architecture by conditioning the generator $G:\mathbb{R}^k \times \mathbb{R}^m \rightarrow \mathbb{R}^n, (z , y) \mapsto G(z|y)$ and discriminator $D:\mathbb{R}^n \times \mathbb{R}^m \rightarrow [0,1], (x , y) \mapsto D(x|y)$, in our case of solving linear inverse problems, on the compressed signal $y$. Therefore, the generative model for $x$ is created not only on the noise $\eta\in\mathbb{R}^m$ injected in the latent space, but also on the particular induced measurement information $y$ which includes implicitly the information of the measurement matrix $A$, making the model more informative. However, this comes at the cost of the adaptability of the network, i.e. changing measurement matrices would require training a new network even if $m$ stays the same.

Besides training GANs in an adversarial manner, auto-encoders can also be trained as a generative model. In this case, the discriminator is trained as an encoder to produce a vector in latent space and the generator as a decoder to produce structured data from a sampled vector in latent space. This class of generative models is known as a Variational Auto-Encoder (VAE). As an example, VAEs in the inverse problem setting have been studied by Bora et al. \cite{pmlr-v70-bora17a}.

It is important to point out that the VAE is trained to minimize the $\ell_2$-loss, the same metric used for measuring reconstruction performance, thus resulting in relatively lower reconstruction losses when recovering compressed measurements. In contrast, GANs are commonly trained to minimize the binary cross entropy loss. How the binary cross entropy loss translates to reconstruction performance is not clear. Moreover, adversarial training bears many difficulties, such as accurate balancing of generator and discriminator loss, hyperparameter selection for training stability, or enforcing diversity on the generated images. To encounter these problems, the so called BEGAN \cite{BEGAN} training method can be applied. In BEGAN, the discriminator is trained as an auto-encoder minimizing the loss function $\| x-D(x) \|_p$, $p\in\mathbb{N}$.

\subsection{Minimax GAN Training }
The minimax method is the original GAN training procedure.
For training GANs, $D$ is trained for maximizing the probability that the labels are correctly assigned to the training examples and samples from $G$. Therefore the following problem is optimized \cite{DBLP:journals/corr/MirzaO14}, directly consider the training of conditional GANs:

\begin{equation}\label{eq:minMax}
\begin{split}
     \min_{G} \max_{D} \mathbb{E}_{x \sim p_x, \eta \sim p_\eta} [\log (D(x|Ax+\eta)] \\+ \mathbb{E}_{z \sim p_z, \eta \sim p_\eta}[1-\log D(G(z|Ax+\eta))].
\end{split}
\end{equation}

The GAN training procedure tries to match the data distribution with the generative distribution. When they both are very similar, this results in a discriminator which is unable to distinguish between real and generated data. One difficulty in the training lies in balancing between the generator and discriminator performance such that both are equally strong, so that an improvement can be made, since e.g. a discriminator that has a very high accuracy would provide relatively uninformative gradients for the generator. This is a necessary condition for the improvement \cite{goodfellow2014generative}. Furthermore, training is done here with respect to a certain noise distribution $p_\eta$. For simplicity, we will consider later training only in the noiseless case $\eta = 0$. It is important to consider that robustness can be improved by injecting some noise.

\subsection{Boundary Equilibrium GAN}
The boundary equilibrium generative adversarial network (BEGAN) \cite{BEGAN} is an alternative GAN training method which considers a different loss function and a different discriminator architecture. Consider again the conditional case for BEGAN. The conditional discriminator has to be adjusted such that it becomes an conditional auto-encoder $D : \mathbb{R}^n \times \mathbb{R}^m\rightarrow \mathbb{R}^n$. The updated loss function $\mathcal{L}_B:\mathbb{R}^n \rightarrow \mathbb{R}^+$ is given by 

\begin{equation}\label{eq:BEGANloss}
    \mathcal{L}_B(x,y) = \|x-D(x|y)\|_2 
\end{equation}
The BEGAN method has the advantage of explicitly keeping the model in the equilibrium between the generator loss and discriminator loss, meaning

\begin{equation}
\begin{aligned}
    &\mathbb{E}_{x\sim p_x, \eta \sim p_\eta}   \left [\mathcal{L}_B(x, Ax + \eta)\right ] \\= &\mathbb{E}_{z\sim p_z, \eta \sim p_\eta}      \left [\mathcal{L}_B(G(z|Ax + \eta),Ax + \eta)\right ].
\end{aligned}
\end{equation}

Introducing $\gamma \in [0,1]$ such that: 
\begin{equation}
    \gamma = \frac{\mathbb{E}_{x\sim p_x, \eta \sim p_\eta} \left [\mathcal{L}_B(x,Ax + \eta)\right ]}{\mathbb{E}_{z\sim p_z, \eta \sim p_\eta}   \left [\mathcal{L}_B(G(z|Ax + \eta),Ax + \eta)\right ]}
\end{equation}
allows us to balance between auto-encoding real image and discriminating real from generated images. Smaller $\gamma$ will lower the image diversity.

The BEGAN training objective considers an autoencoding loss function $\mathcal{L}_B$ and a given equilibrium $\gamma$ which is controlled throughout the training process. The objective consists of minimizing the discriminator loss $\mathcal{L}_D$ and generator loss $\mathcal{L}_G$
\begin{equation} \label{eq:BeganLoss}
    \begin{aligned}
        \mathcal{L}_D(x,y) &= \mathcal{L}_B(x,y) - \beta_n \mathcal{L}_B(G(z|y),y) \\
        \mathcal{L}_G(x,y) &= \mathcal{L}_B(G(z|y),y)\\
    \end{aligned}
\end{equation}
where $\beta_n$ for each training step $n$ is controlled by $\beta_n = \beta_{n-1} + \lambda_n (\gamma \mathcal{L}_B(x,y) - \mathcal{L}_B(G(z|y),y))$ with $\lambda_n$ as the proportional gain or learning rate for $\beta_n$. In the beginning, $\beta_0$ is initialized to 0.

\subsection{Implementing the Conditioning}

\begin{figure*}[t!]
\begin{center}
\includegraphics[width=0.85\textwidth]{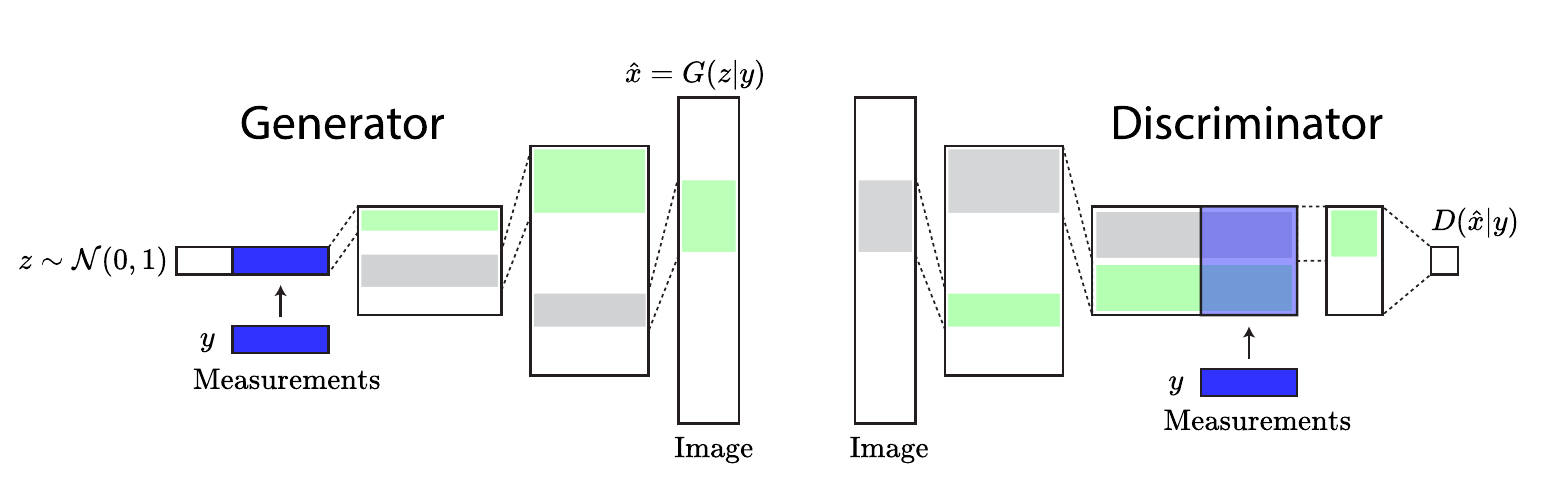}
\caption{\textbf{Inducing Conditioning into the Network} In the generator the conditioning $y$ is concatenated into the latent variable $z$. In the discriminator the image goes through the convolutional and pooling layers before the conditioning is induced, where the measurements is repeated spatially to match the size of the filters on the last layer. Afterwards, the last layer with the induced measurements goes through a 1x1 convolutional layer.}\label{fig:GANConditioning}
\end{center}
\end{figure*}

At training start of GANs, the samples generated by $G$ do not resemble samples coming from the distribution of the dataset and thus rejected by $D$ with high confidence, leading to saturating gradients when minimizing $\mathbb{E}_{z \sim p_{z}, \eta \sim p_\eta}[1-\log D(G(z|Ax + \eta))]$ in (\ref{eq:minMax}). In practice, this optimization for training $G$ can be improved by replacing it by the maximization of $\mathbb{E}_{z \sim p_{z}, \eta \sim p_\eta}[\log D(G(z|Ax+\eta))]$.

How to optimally induce the conditioning into the network is an open question. 
One way to induce conditioning into the network \cite{pmlr-v48-reed16} is illustrated in Fig. \ref{fig:GANConditioning}. For the generator the condition $y$ is concatenated to the latent variable $z$. In order to insert the conditioning into the discriminator, the image is first sent through convolutional layers such that the spatial dimension is reduced to a small size (in our case, $4\times 4$). The conditioning $y$ is then induced by replicating the conditioning $y$ spatially to match the size of the last output and concatenating it across the channel dimension on the last output. Consequently, the concatenated layer goes through a $1\times 1$ convolutional layer reducing the number of channel dimension and passed through a final convolutional layer to return the probability of $x$ being generated. Further improvements can be made by updating the loss function (Algorithm \ref{cGAN}), where the discriminator loss also evaluates the prediction given true values.   

Due to the conditioning, the generative model is more granular for each target signal compared to its unconditional counterpart. In compressed sensing tasks, this increases the chance of finding $x^*$. This should result in a smaller error and a more realistic reconstruction \cite{kim2020compressed}. The conditioning on $y$ can be combined with various reconstruction methods. The training algorithm for conditional GANs can be derived from standard GAN \cite{goodfellow2014generative} by adding the measurements to the inputs of the generator and discriminator.

\begin{algorithm}\label{cGAN}
 \caption{Conditional GAN Training Algorithm \cite{pmlr-v48-reed16}}
    \SetKwFunction{Conditional GAN}{isOddNumber}
    \SetKwInOut{KwIn}{Input}
    \SetKwInOut{KwOut}{Output}

    \KwIn{$x$: Minibatch of Images, $A$: Measurement Matrix, $D_{\phi}$: Discriminator with Weights $\phi$, $G_{\theta}$: Generator with Weights $\theta$, $N$: Number of Batches to be Processed, $\alpha$: Step Size}

  \While{$n<N$}{
    $\eta  \sim p_\eta$ \\
   $ y= Ax + \eta$\\
  $ z \sim \mathcal{N}(0,I_k) $\\
   $\hat{x} = G_{\theta}(z|y)$ \\
   $\mathcal{L}_D = (\log D_{\phi}(x|y)  + \log(1-D_{\phi}(\hat{x}|y)))/2$ \\   
   $\phi = \phi - \alpha\nabla \mathcal{L}_D $ \\
   $\mathcal{L}_G = \log (D_{\phi}(\hat{x}| y)) $\\
   $\theta = \theta - \alpha \nabla \mathcal{L}_G $
   }
\end{algorithm}

\subsection{Comparing to Variational Autoencoders}
GANs utilize a generator to map from latent space to image space and a discriminator to evaluate whether the image was generated of not. In the case of (unconditional) auto-encoders, one trains an encoder which compresses images to latent space and a decoder that again maps latent to images space. In this context, a natural loss function is given by the reconstruction error. Thus, the decoder acts as an (unconditional) generator $ G:\mathbb{R}^k \rightarrow \mathbb{R}^n$. In this work we restrict to unconditional auto-encoders, since inducing conditioning to auto-encoders is also a possibility. In the case of Variational Autoencoders (VAE) the latent space is assumed stochastic. The encoder learns a mapping from image space to the latent space which is parameterized by the mean and variance of some distribution, in particular usually a Gaussian distribution $q_{\theta}(z|x)=\mathcal{N}(z; \mu, \sigma^2 I_n)$. The decoder learns to map the stochastic latent space to some distribution $p_{\phi}(x|z)$ of the image space. The model is then trained by maximizing the variational lower bound given by
\begin{equation*}
    \argmax_{\phi,\theta} -\mathbb{E}_{z\sim q_{\phi}(z|x)}\left [ \log p_{\theta}(x|z) \right ] + \mathbb{KL}(q_{\phi}(z|x)\| p(z)),
\end{equation*}
where $\mathbb{KL}$ then denotes the Kullback-Leibler divergence. The decoder can be used to generate images which are similar to the training data. A sample can be produced by sampling a latent variable $z\sim \mathcal{N}(\mu, \sigma^2 I_k)$. Subsequently, the sample is passed to the generator to produce $x=G(z)$.


\section{Compressed Sensing using Generative Models}\label{sec:CSuGAN}

Various methods have been proposed for using neural networks (NNs) to solve compressed sensing problems. One active field of research is, for instance algorithm unrolling \cite{monga}, where a classical iterative reconstruction algorithm is unrolled/unfolded into a multilayer perceptron network. One of the first work here is LISTA \cite{LISTA}. Another direction, which is also our focus, uses NNs as generative priors for compressed sensing recovery. It is shown that conditional generative models can achieve a better performance than its unconditional counterpart by implicitly including information of the sensing device given by the measurement matrix \cite{kim2020compressed}.

Several NN architectures can be employed, such as VAEs or GANs trained using the standard min-max strategy and the BEGAN procedure. For the latent space, it is assumed that for every $x^*$ there exists a vector $\hat{x} = G(z^*|y)$ that is very close to $x^*$. The approximated solution $\hat{x}$ to the inverse problem (\ref{eq:INV}) is estimated by solving the optimization problem 
\begin{equation}\label{eq:TestCSGM}
    z^* = \argmin_{z} \| y - A G(z|y) \|_2^2.
\end{equation}
In this case, the loss function $\mathcal{L}$ in \eqref{eq:loss_equation} is the squared $\ell_2$-norm.

\subsection{Projected Gradient Descent}
A prominent method for solving (\ref{eq:loss_equation}) or (\ref{eq:TestCSGM}) is projected gradient descent (PGD).
The difference to standard gradient descent is that PGD is performed in the ambient space $\mathbb{R}^n$. 
The optimization consists of two steps. First, gradient descent is applied to $\| y - Ax\|^2_2$ which results in the gradient descent update of the estimate $w_n$ at the $n$-th iteration, 
\begin{equation}\label{eq:PDG1}
    w_n = x_n + \mu A^T(y - Ax_n), 
\end{equation}
where $\mu$ is the step size for $n = 0, ..., N-1$, with $N$ being the number of iterations. In general, the initial condition is set as $x_0 = 0$ or $x_0 = A^T y$. Afterwards in the projection step, the projection $P_G$ tries to find the closest point to $w_n$ in the image of $G(\cdot|y)$:
\begin{equation}\label{eq:PDG2}
    \mathcal{P}_G(w_n|y) = G(\argmin_{z} \| w_n - G(z|y)\|_2|y) 
\end{equation}
\begin{algorithm}
 \caption{Projected Gradient Descent}
    \SetKwFunction{Projected Gradient Descent}{isOddNumber}
    \SetKwInOut{KwIn}{Input}
    \SetKwInOut{KwOut}{Output}

    \KwIn{ $y$, $A$, $G$, $N$}
    \KwOut{$\hat{x}$}

   $x_0 = 0$ \\
   \While{$n<N$}{
   $ w_n= w_n + \eta A^T(y-Ax_n)$\\
   $x_{n+1} =   \mathcal{P}_G(w_n|y) = G(\argmin_{z} \| w_n - G(z|y)\|_2|y) $\\
   $n = n+1$\\
   }
   $\hat{x}=x_N$
\end{algorithm}

A fundamental question is, under which condition meaningful recovery is possible, i.e. can be ensured. At the moment, the research is here at the very beginning and only generic results are available. For example, in \cite{shah2018solving}, the case of Gaussian measurement matrices is considered and guarantees can be established which hold with high probability. As in the history of compressed sensing, this is a common approach to develop more theory. To state a result in this direction we recall the definition of the S-REC condition \cite{shah2018solving}.
The $\text{S-REC}(\mathcal{S}, \gamma, \delta)$ condition for a matrix $A$ and scalars $\gamma >0$, $\delta\geq 0$ is satisfied when 
\begin{equation}
    \| A(x_1 - x_2)\|_2^2 \geq \gamma \|x_1 - x_2\|_2^2 - \delta
\end{equation}
holds for all $x_1,x_2 \in \mathcal{S}$.

Now, the image of a smooth generator is a $k$-dimensional manifold in ambient dimension $n$ and based on results in \cite{shah:allerton11}, the following convergence result has been established by Shah and Hedge.\\

\textbf{Theorem 1:} [Theorem 2.2 in \cite{shah2018solving}] \emph{Let $G$ be a differentiable generator and $A$ a random Gaussian matrix with $A_{ij}\sim \mathcal{N}(0,1/m)$ satisfying the $\text{S-REC}(\mathcal{S}, \gamma, \delta)$ with probability $1-p$ and $\|Ax \|_2 \leq \rho \|x\|_2$ for all $x\in \mathbb{R}^n$ with probability $1-q$, where $\rho^2\leq \gamma $, then the sequence $(x_n)_{n\in \mathbb{N}}$ defined by PGD for $y= Ax^*$ will converge to $x^*$ for every $x^* \in \mathcal{S}$ with probability at least $1-p-q$.}\\  

Hence, to some extent $\text{S-REC}(\mathcal{S}, \gamma, \delta)$ together with the upper bound $\rho$ in the theorem play a similar role as the RIP condition \eqref{eq:rip} and ensuring recovery with high probability (here in the noiseless case).

\subsection{Network-Based PGD}\label{sec:NPGD}
In order to improve computation time and target a possible real-time reconstruction, a carefully constructed NN can
be employed to replace the inner loop in the PGD
(\ref{eq:PDG2}) and mimic the task of a pseudo-inverse. Such an approach substantially speeds
up the overall reconstruction. The disadvantage of this so called
``network-based PGD'' approach (NPGD) is that a new pseudo-inverse
network must be created first, and hence is more suitable for frequent
decoding of compressed signals of the same setup and architecture.

For this purpose we target a map $P_S$ which is an ``approximate
projector'', i.e., it should satisfies: (1) for all $x\in \mathbb{R}^n$ $(P_S \circ P_S)(x) \approx P_S(x)$ and (2) $P_S(\Tilde{x}) \approx \argmin_{x\in S} \|\Tilde{x} - x\|_2^2$ \cite{Raj_2019_ICCV}. The operator $G_\theta^+:\mathbb{R}^n \times \mathbb{R}^k \rightarrow \mathbb{R}^k$ is an approximate pseudo-inverse of $G$. Due to $G$ being non-linear a NN can be used to create $G_\theta^+$.

The network parameters for the approximate pseudo-inverse are computed by 
\begin{equation}\label{eq:inverse_loss}
\begin{aligned}
     \min_{\theta} 
     \mathbb{E}_{\begin{subarray}{l}z , \nu , y \end{subarray} }[ &\|G(G_\theta^{+}(G(z|y) + \nu |y)|y) - G(z|y)\|^2_2  \\ + &\lambda \|(G_\theta^{+}(G(z|y) + \nu) - z|y)\|^2_2 ],
\end{aligned}
\end{equation}
where $z \sim p_z=\mathcal{N}(0,I_k)$, $\nu \sim p_{\nu}=\mathcal{N}(0,I_m)$ and $y = Ax +\eta $ with
$\eta \sim p_{\eta} = \mathcal{N}(0,\sigma^2 I_m)$ an additive
Gaussian noise as in \eqref{eq:INV}.  
Keeping $G$ fixed and minimizing the loss (\ref{eq:inverse_loss}) with respect to $\theta$ results in the optimization of $G^+$.  The projector $\mathcal{P}_G $ introduced for PGD is then replaced to $\mathcal{P}_G = G \circ G^+_\theta$.
In the experiments, an
unconditional pseudo-inverse $G^+$ is used for simplicity, but this
could be easily  extended to the conditional case leading to possibly further improvements.

In NPGD, the first optimization step is (\ref{eq:inverse_loss}) is same as with the PGD case. The second step removes the inner loop by computing directly 
\begin{equation}
    x_{n+1} = G(G^+(w_n|y)|y).
\end{equation}
\begin{algorithm}\label{NPGD_Algorithm}
 \caption{Network-Based Projected Gradient Descent}
    \SetKwFunction{Projected Gradient Descent}{isOddNumber}
    \SetKwInOut{KwIn}{Input}
    \SetKwInOut{KwOut}{Output}

    \KwIn{  $y$, $A$, $G$, $N$, $G^+$}
    \KwOut{$\hat{x}$}

   $x_0 = A^Ty$ or $x_0=0$ \\
   \While{$n<N$}{
   $ w_n= w_n + \mu A^T(y-Ax_n)$\\
   $x_{n+1} =   G(G^+(w_n|y)|y)$\\
   $n = n+1$\\
   }
   $\hat{x} = x_N$\\
 \end{algorithm}
 First theoretical results exists for NPGD \cite{Raj_2019_ICCV} and e.g.
 are based on the notion of the restricted eigenvalue constraint (REC) and $\delta$-approximate projector. \\
More precisely, 
the restricted eigenvalue constraint
$\text{REC}(\mathcal{S},\alpha,\beta)$ with respect to  a set
$\mathcal{S}\subset \mathbb{R}^n$ and parameters $0<\alpha < \beta$ is satisfied for a matrix $A\in\mathbb{R}^{m\times n}$ , if the following condition holds $\forall x_1,x_2\in \mathcal{S}$
\begin{equation}
    \alpha\| x_1 - x_2\|^2_2\le \|A(x_1-x_2)\|^2_2 \le \beta\|x_1-x_2\|^2_2.
\end{equation}
For clear presentation we refer here to the original formulation \cite{Raj_2019_ICCV} for unconditional generators.
A composite network $G \circ G^+ : \mathbb{R}^n \to  \mathbb{R}^n$ is a $\delta$-approximate projector, if there exist a $\delta > 0$ such that $\forall x\in\mathbb{R}^n$ 
\begin{equation}
    \| x - G(G^+(x)) \|_2^2 \leq \min_{z\in \mathbb{R}^k} \| x - G(z) \|^2_2 + \delta
\end{equation}
holds. Now, let $f=\|Ax-y\|_2^2$ be the loss function of PGD.\\

\noindent \textbf{Theorem 2:} [Theorem 1 in \cite{Raj_2019_ICCV}]
\emph{Let $A\in \mathbb{R}^{m\times n}$ be a matrix that satisfies
  $\text{REC}(\mathcal{S}, \alpha, \beta)$ with $\beta / \alpha < 2 $,
  and let $G \circ G^+$ be a  $\delta$-approximate projector. Then for
  every $x^*\in\text{range}(G)\subset\mathbb{R}^n$ and measurement $y
  = Ax^*$, executing Algorithm \ref{NPGD_Algorithm} with step size
  $\eta=1/\beta$, will yield $f(x_n ) \leq (\frac{\beta}{\alpha} -1)^n
  f(x_0) + \frac{\beta \delta}{2 -
    \frac{\beta}{\alpha}}$. Furthermore, the algorithm achieves $\|
  x_n-x^*\|^2_2 \leq (C+ \frac{1}{2\frac{\alpha}{\beta}- 1})\delta$
  after
  $\frac{1}{2-\frac{\beta}{\alpha}}\log(\frac{f(x_0)}{C\alpha\delta})$
  steps. When $n\rightarrow\infty$, $\|x^*-x_{\infty}\|^2_2 \le \frac{\delta}{2\alpha/\beta -1}$ }.

Thus, linear convergence is guaranteed if the matrix has a sufficiently good
$\text{REC}$ property. However, as the discussion in
\cite{Raj_2019_ICCV} shows, common (unstructured) random matrices fail to
satisfy $\beta/\alpha<2$ with sufficiently high probability and
several approaches are discussed to find measurement matrices with
almost-orthogonal rows. 

The projection error of $G \circ G^+$ will be visible in the results.

\section{Experiments}
In this section we demonstrate the proposed architecture for different setups.
For the experiments we have used a batch of $C$ images for which the different algorithm discussed in Section \ref{sec:CSuGAN} are implemented and the images are recovered from compressed measurements. 

In the following experiments, for the compressed image $Ax$ and an additive noise $\eta \in \mathbb{R}^{m}$ the signal to noise ratio (SNR) is defined by
\begin{equation}\label{eq:SNR}
    SNR = \frac{\|Ax\|_2^2}{ \|\eta\|_2^2}.
\end{equation}
The mean squared error (MSE) is defined by 
\begin{equation}
\text{MSE} = \frac{1}{C}\sum_{i=1}^{C} \|x_i - x^*_i\|_2^2  .   
\end{equation}
The residual error is defined as 
\begin{equation}
     \frac{1}{C} \sum_{i=1}^{C}   \|Ax_i - y_i\|_2^2
\end{equation}
The structural similarity index is defined as 
\begin{equation}
    \text{SSIM}(x,x^*) =\frac{(2\mu_x \mu_{x^*} + C_1)(2\sigma_{x x^*} +C_2)}{(\mu_x + \mu_{x^*} + C_1)(\sigma_x + \sigma_{x^*} + C_2)},
\end{equation}
where $\mu_x, \mu_{x^*}$ are the estimated means and $\sigma_{x x^*}, \sigma_{x },\sigma_{x^*}$ are the estimated covariance and variances.
The mean SSIM \cite{SSIM} uses an $M$-sized window to evaluate the SSIM
\begin{equation}
    \text{MSSIM} =  \frac{1}{M} \sum_{i=1}^M \text{SSIM}(x_i,y_j).
\end{equation}
SSIM is a perception-based measure for similarity of two images. It takes values between $0$ and $1$ where SSIM=$1$ can only be reached when both images are the same. Compared to MSE, SSIM better mimics aspects on human perception by capturing interdependencies between pixels, revealing important information about an object in the visual scene \cite{SSIM}. In various fields such as image decompression, image restoration, or pattern recognition, SSIM often serves as a more reasonable metric than the MSE.

\subsection{Experiments and Results for MNIST}
For the first experiments the MNIST dataset was used.
The dataset was selected due to the desired properties such as its smaller size and lower structural complexity compared to other datasets, while giving simple insights into the general attributes of the different networks. 
A batch size of $C = 64$ images for reconstruction was chosen. 
The MNIST dataset consists of 70.000 greyscale images of handwritten digits of size $28\times 28$. The training set is composed of 60.000 images and the test set consists of 10.000 images \cite{MNIST}.

The measurement matrix was normalized based on the number of measurement taken $m$ resulting in $A_{i,j} \sim \mathcal{N}(0, 1/m)$. 

\begin{description}
    \item[PGD Hyperparameters] The number of inner and outer iterations are chosen to be 100 and 30,  respectively, to achieve a balance between optimization time and accuracy. The learning rate is chosen to be 0.01 for the inner loop and 0.5 for the outer loop. These rates are based on the implemented values by Shah et al. \cite{shah2018solving}. The initial value is $\hat{x}_0 = 0$.
    
    \item[NPGD Hyperparameters] The number of outer iterations is kept at $N = 30$ for a fair comparison to the PGD procedure, however further speedup could be produced with avoiding as many outer loop iterations without any performance loss (\ref{NPGD_Algorithm}). From our experience, with $3$ to $5$ outer loop iterations convergence of the error is already achieved and further iterations would not provide significant improvements.

    \item[Training procedures] We compared three different generative model training procedure, the standard minimax GAN procedure \cite{goodfellow2014generative}, BEGAN procedure \cite{BEGAN}, and a variational procedure \cite{kingma2013auto} as a comparison. For GANs, we always consider the standard GAN training procedure unless stated otherwise.  The measurement-conditional models where trained for the noiseless measurement conditioning $y=Ax$. The models were trained for 200 epochs using Adam optimizer ($\beta_1=0.5,\beta_2=0.999$) and learning rate $0.0001$. 
        
    \item[Model Architectures] A multilayer perceptron GAN (MLPGAN), a deep convolutional GAN (DCGAN), a measurement-conditional DCGAN (mcDCGAN), and a measurement-conditional DCGAN trained with the BEGAN procedure (mcDCBEGAN) were used in the experiment. A variational autoencoder (VAE) architecture was used as an comparison.  
    
    \item[Pseudo-Inverse Generator] The architecture of the pseudo-inverse generator network is the same as that of the discriminator, except that the scalar output layer is replaced by a fully-connected linear layer of size $k$.  This approach was already successfully applied in \cite{invCond}.  In particular, we trained the inverse-generator network for 100 epochs and set $\lambda=0.1$ and $\sigma^2=1.0$.
\end{description}

\begin{center}
    \textbf{Quantitative Results}
\end{center}

\begin{figure}[H]
    \centering
    Subsampling Ratio 2\%\\
    \includegraphics[width=1\textwidth]{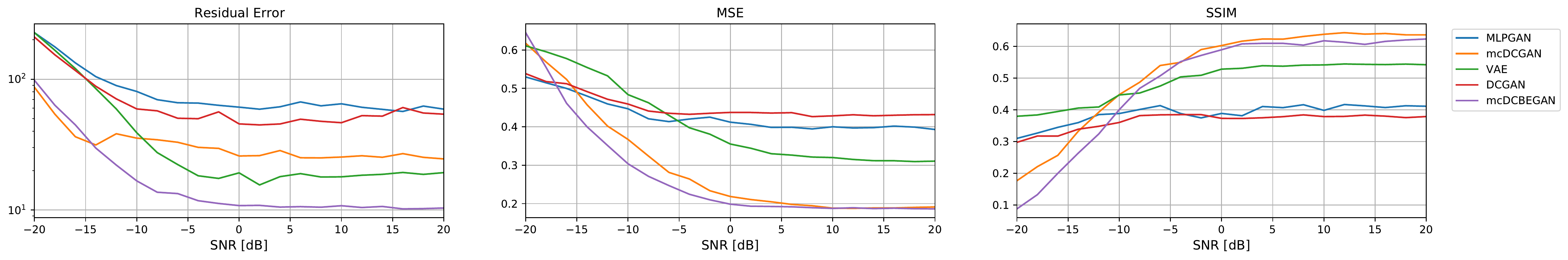}\\
    Subsampling Ratio 5\%\\
    \includegraphics[width=1\textwidth]{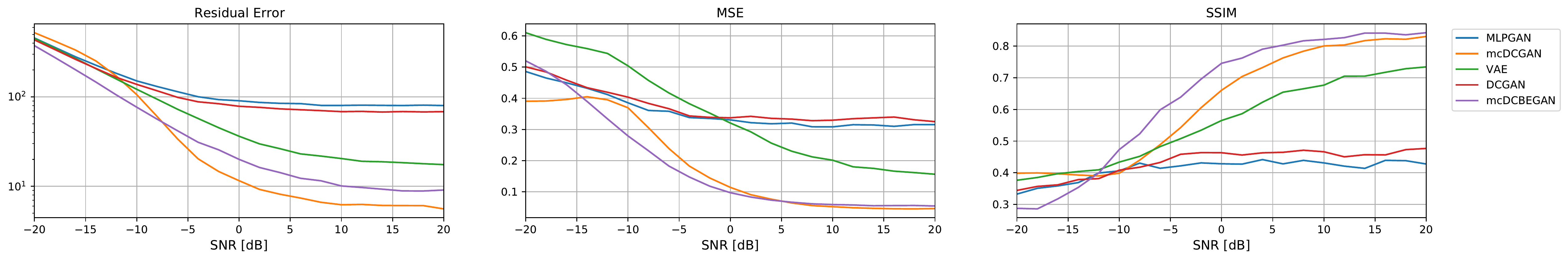}\\
    Subsampling Ratio 10\%\\
    \includegraphics[width=1\textwidth]{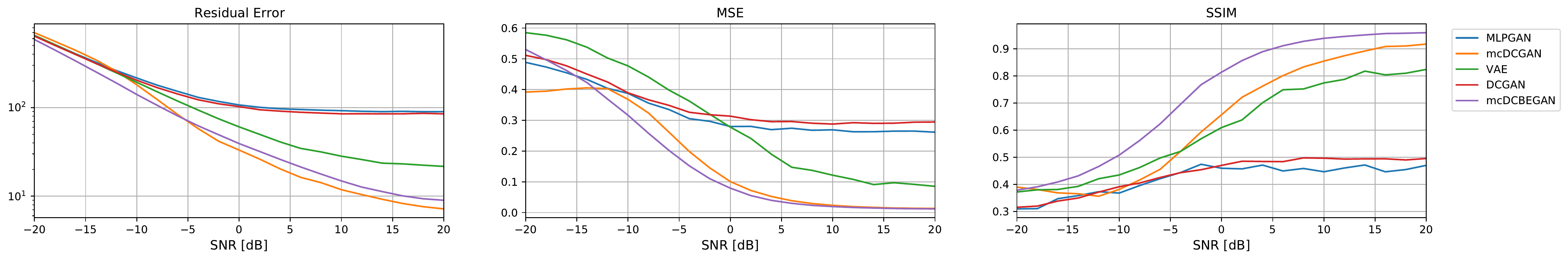}\\
    \caption{\textbf{MNIST Reconstruction Residuals using PDG}. Initialization with $\hat{x}_0 = 0$. SNR ranges from $-20$dB to $20$dB in $2$dB steps according to (\ref{eq:SNR}). As expected the conditioning on the measurement improves the MSE, specifically at a factor of $\sim3$ for a $2\%$ sub-sampling rate and at a factor $\sim 6$ for a sub-sampling rate of $5\%$. The BEGAN architecture performs equally to the mcDCGAN for SNR larger than 0. Until the crossing at around $-10$dB the mcDEBEGAN architecture is superior, due to begin implicitly trained on the image reconstruction (\ref{eq:BeganLoss}). The SSIM shows the perceived closeness of images, specifically higher values mean higher perceived similarity. The better performance of measurement conditional model is due to the conditioning on the latent space which increases certainty.}
    \label{fig:MNIST_PDG_LOSSES}
\end{figure}

The conditioning on the measurements is shown to produce a clear advantage compared across all losses. Comparing the MSE in Fig. \ref{fig:MNIST_PDG_LOSSES}, the conditional model produces results which are twice as good. This improvement is achieved by the reduction of uncertainty on $x$ by introducing the conditioning variable. 

\begin{figure}[H]
    \centering
    Subsampling Ratio 2\%\\
    \includegraphics[width=1\textwidth]{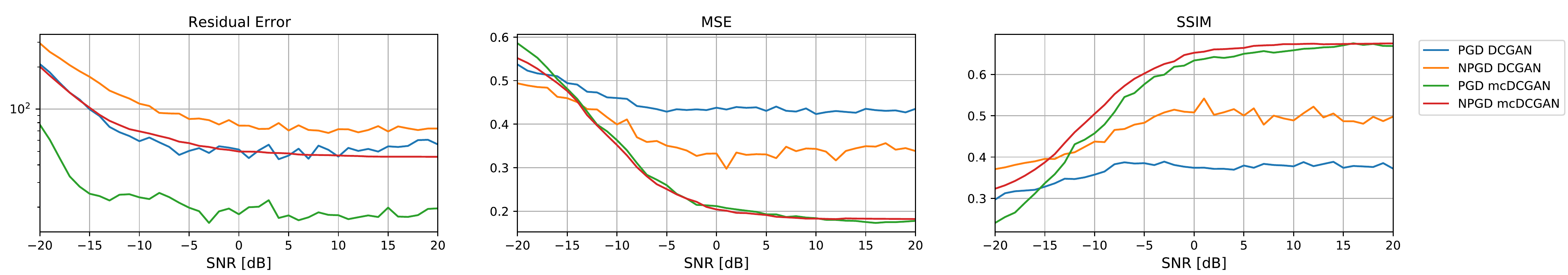}\\
    Subsampling Ratio 5\%\\
    \includegraphics[width=1\textwidth]{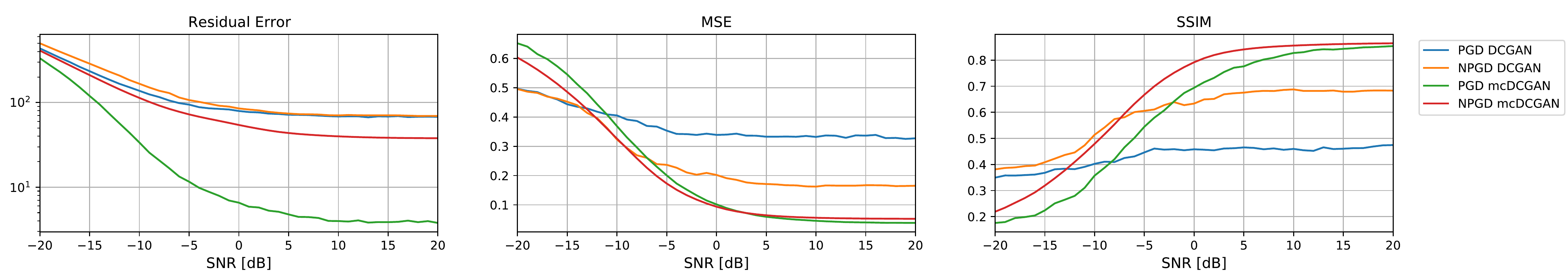}\\
    Subsampling Ratio 10\%\\
    \includegraphics[width=1\textwidth]{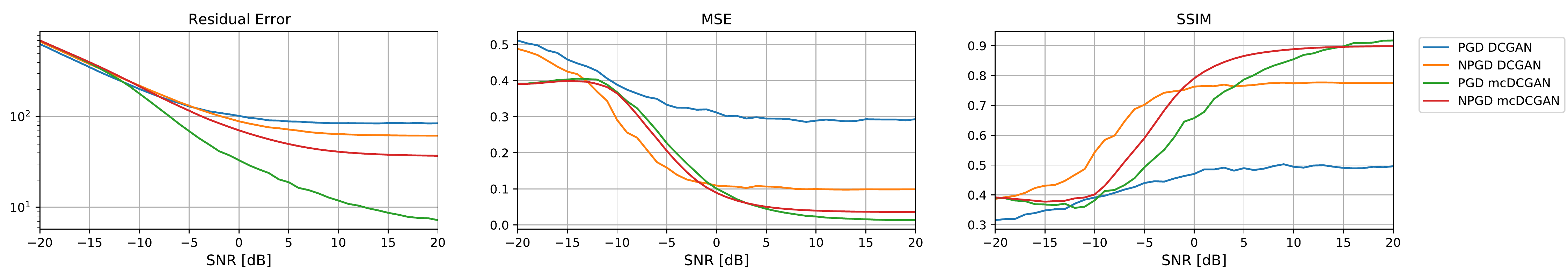}\\

    \caption{\textbf{MNIST Reconstruction Residuals using NPDG and PGD.} The performance difference between PGD and NPGD is dependent on using an measurement conditional or unconditional GAN. For the unconditional GAN the NPGD is able to improve the results a lot, this could be due to PGD getting stuck in a local minimum. The pseudo-inverse generator for the NPGD is learning the ambient space which then produce results that are much closer to the original value $x^*$, without having to search the ambient space with gradient decent (\ref{eq:PDG1}). The smaller differences between PGD and NPGD for the conditional models could be due to a closer reconstruction to the original value since the conditioning is increasing the certainty. Therefore the probability of PGD getting stuck in a local minima is much smaller. The projection error of NPGD is visible in the conditioned models for high SNR, but the implementation of the NPGD algorithm was only done to speedup the recovery. In our experience the NPGD speedup is in a range of $140-175$x. Similar results are also visible for a MLPGAN in Fig. \ref{fig:mlp_losses_m39}.}
    \label{fig:MNIST_NPDG_MLP_DCGAN}
    
\end{figure}

In the next phase, the pseudo-inverse network $G^+$ (Section \ref{sec:NPGD}) was introduced as a replacement of the inner loop.  For the unconditional model, the pseudo-inverse network is better able to find the a local minimum. We observed an improvement of 2-3$\times$ better reconstruction MSE. The results showed similar improvements when using an MLPGAN. The NPGD is in our experience about 140-175x faster than PGD. For the conditional model reconstruction performance is approximately 20-30 \% worse (Fig. \ref{fig:MNIST_NPDG_MLP_DCGAN}) however reconstruction was completed at a much faster rate. This results in a trade-off between the reconstruction time and the accuracy. Further possible improvements could be made by conditioning the pseudo-inverse $G^+$. 

\begin{figure}[!t]
    \centering
    \includegraphics[width=0.45\textwidth]{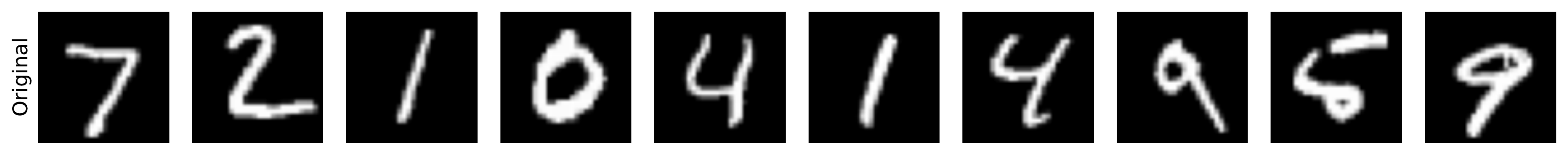}\\
    \small Unconditional Generative Model\\
    \tiny Subsampling Ratio 10\%, Noiseless Case\\(Best case for reconstruction)\\
    \includegraphics[width=0.45\textwidth]{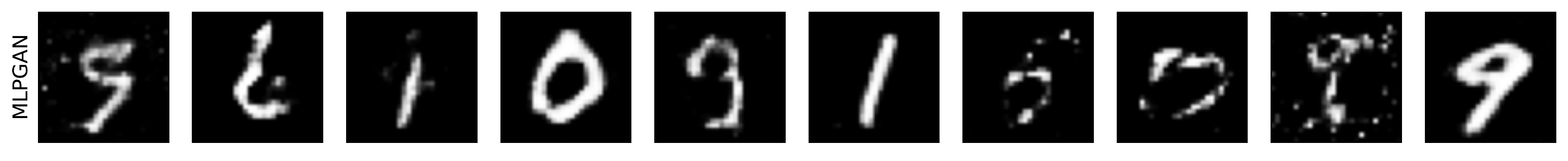}\\
    \includegraphics[width=0.45\textwidth]{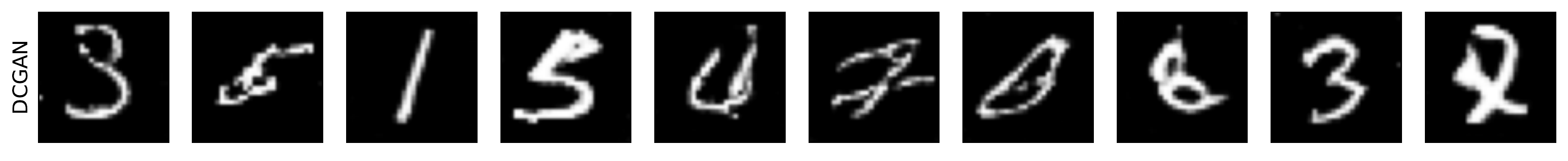}\\
    \includegraphics[width=0.45\textwidth]{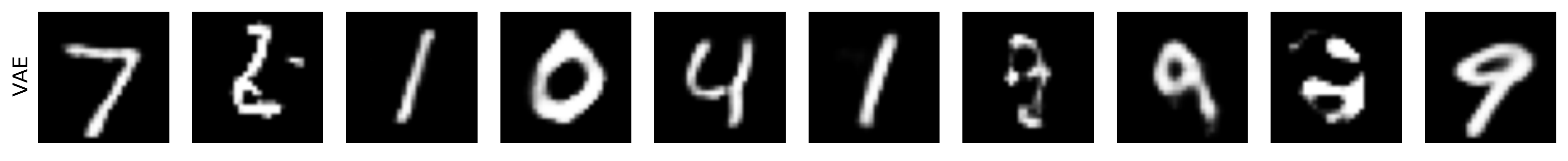}\\
    
    \small Measurement-conditioned GAN\\
    \tiny Subsampling Ratio 2\%, SNR 0dB \\
    \includegraphics[width=0.45\textwidth]{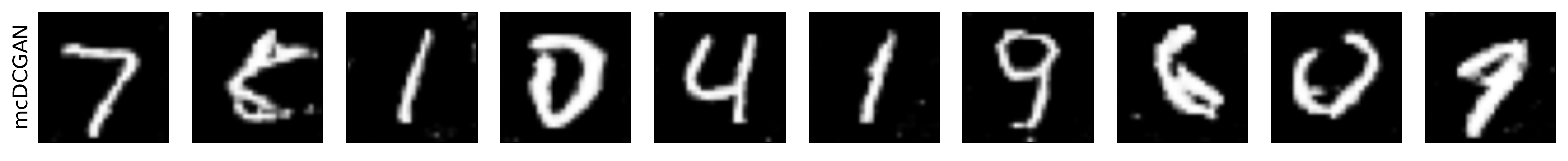}\\
    \includegraphics[width=0.45\textwidth]{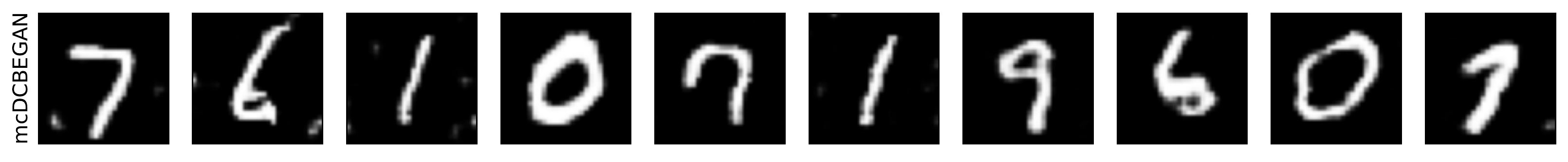}\\
    \centering
    \tiny Subsampling Ratio 2\%,  Noiseless Case\\
    \includegraphics[width=0.45\textwidth]{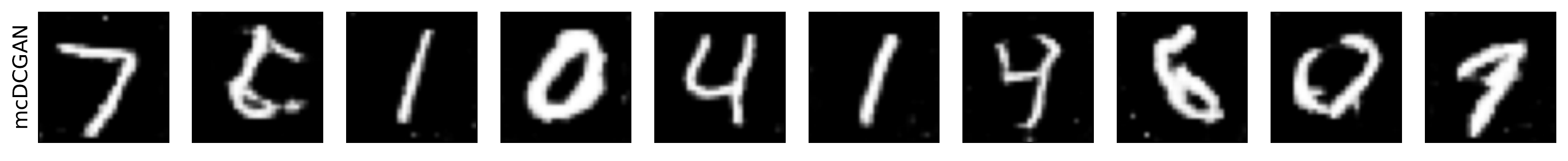}\\
    \includegraphics[width=0.45\textwidth]{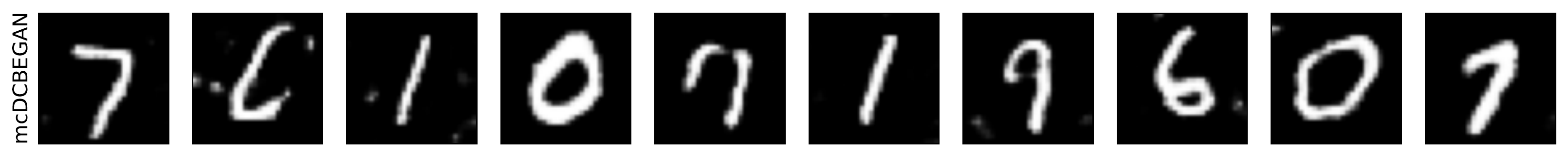}\\
    \tiny Subsampling Ratio 5\%, SNR 0dB \\
    \includegraphics[width=0.45\textwidth]{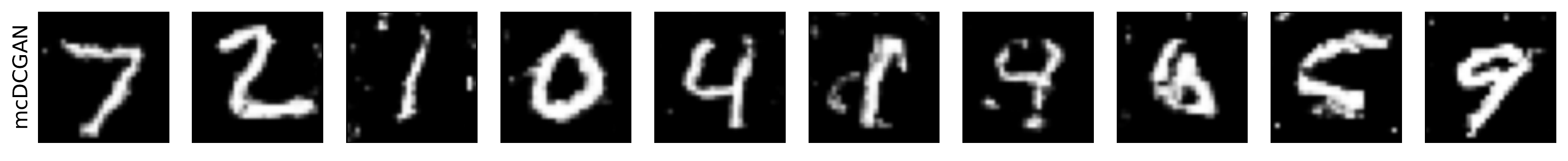}\\
    \includegraphics[width=0.45\textwidth]{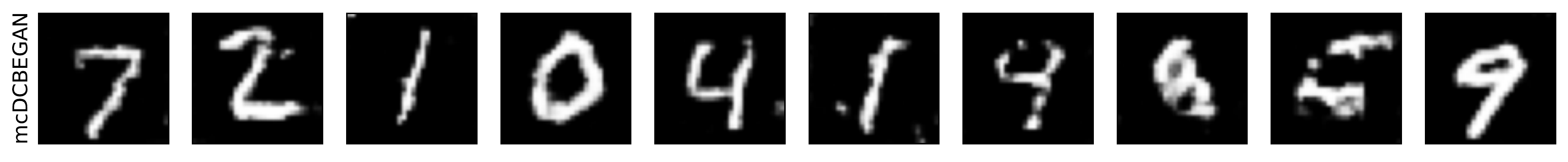}\\
    \tiny Subsampling Ratio 5\%,  Noiseless Case\\
    \includegraphics[width=0.45\textwidth]{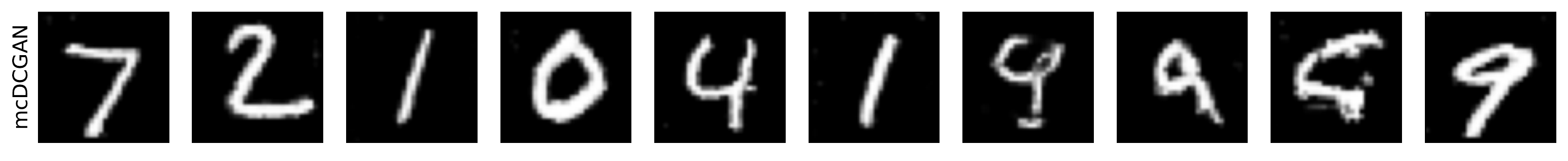}\\
    \includegraphics[width=0.45\textwidth]{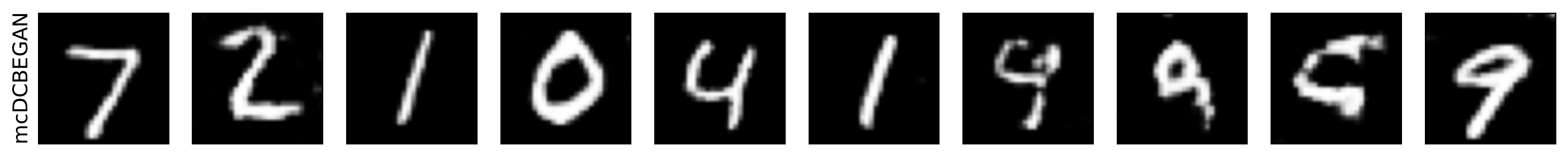}\\
    \tiny Subsampling Ratio 10\%, SNR 0dB \\
    \includegraphics[width=0.45\textwidth]{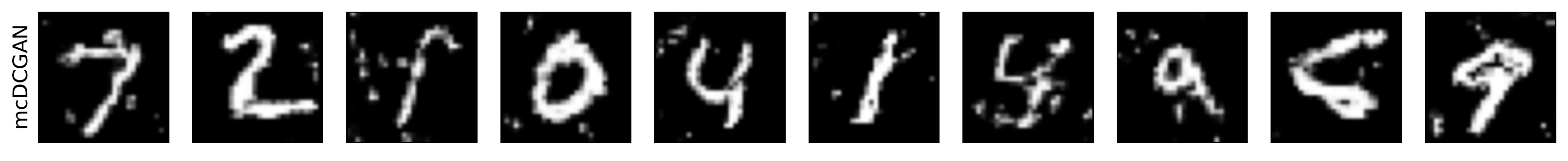}\\
    \includegraphics[width=0.45\textwidth]{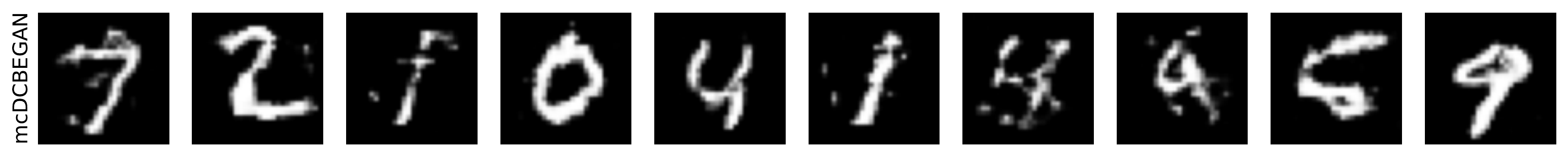}\\
    \tiny Subsampling Ratio 10\%, Noiseless Case\\
    \includegraphics[width=0.45\textwidth]{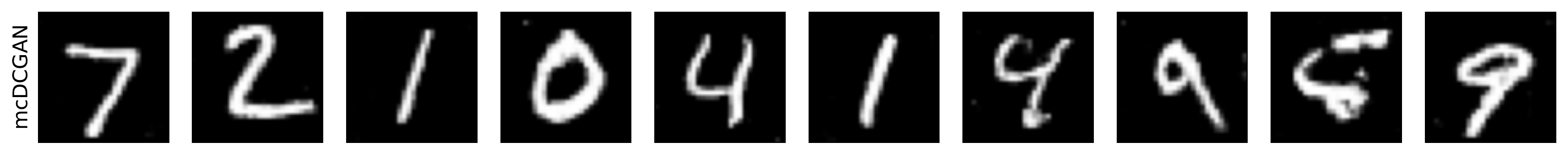}\\
    \includegraphics[width=0.45\textwidth]{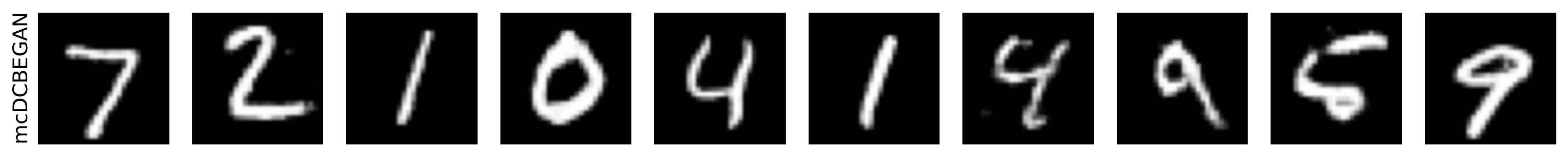}\\
    \caption{\textbf{MNIST Reconstructed Images from PDG} The unconditional models reconstruct the compressed numbers at much lower accuracy. Conditioning is increasing the reconstruction performance, the $0$dB for 5\% and 10\% most numbers are reconstructed correctly, while in the noiseless case all numbers are reconstructed really close to the original signal. In the 2\% case mcDCGAN is reconstructing the 4 while mcDCBEGAN is not, but evaluating at the MSE (Fig. \ref{fig:MNIST_PDG_LOSSES}) the mcDCBEGAN is performing better at $0$dB. The classification we infer when looking at the images is not of intrest in compressed sensing.}
    \label{fig:MNIST_PDG_IMAGES}
\end{figure}

\vspace{5cm}

\subsection{Experiments and Results for CelebA}
Since the MNIST dataset does not capture the sophisticated structure of natural images, the experiment in the previous subsection is repeated to the CelebA dataset, which is a much more complex dataset compared to MNIST. The CelebA dataset consists of 202599 samples. For reconstruction a batch size of 128 is chosen. 
\begin{description}
\item[PGD Hyperparameters] The number of inner and outer iterations are chosen to be 100 and 30,  respectively. The learning rate is chosen to be 0.01 for the inner loop and 0.5 for the outer loop.
\item[Training Procedures] As with the MNIST dataset, we compared three different generative model training procedure, the standard minimax GAN procedure and BEGAN procedure. The dataset was split into 160000 samples for training and 42599 for evaluation. For training the DCGAN we used a learning rate of $0.0002$.
\item[Model Architectures] Three model architectures were chosen, in particular a deep convolutional GAN (DCGAN), a measurement-conditional DCGAN (mcDCGAN), and a measurement-conditional DCGAN trained with the BEGAN procedure (mDCBEGAN).
\item[Pseudo-Inverse Generator] The architecture of the pseudo-inverse generator network is the same as that of the discriminator, except that the scalar output layer is replaced by a fully-connected linear layer of size $k$.
\end{description}

\vspace{1cm}
\begin{center}
    \textbf{Quantitative Results}
\end{center}
The advantage of using measurement-conditional models is not clear when applied to the CelebA dataset in contrast to the MNIST dataset. There are two major possible factors that reduce the reconstruction ability of the measurement-conditional models: 1. the models are only trained for fewer epochs compared to MNIST since the architecture is more complex hence training takes more computation time, and 2. the conditioning on $y$ is implemented for the noiseless case and not in the noisy case, since there are some possible improvements that could be made from using some noise during the training of the model. 

In the low SNR case, the NPGD has shown to perform worse compared to using PGD for the unconditional model, however this performance gap is much smaller for the measurement conditional model. The NPGD always performs worse than the PGD, since there is a trade-off between the speed and accuracy. The combination of the GAN architecture with the autoencoder loss (\ref{eq:BEGANloss}) for the discriminator allows the mcDCBEGAN network to better reconstruct images for a SNR value of $-5$ to $0$dB.

\onecolumngrid

\begin{figure*}[h!]
\begin{center}
\includegraphics[width=.8\textwidth]{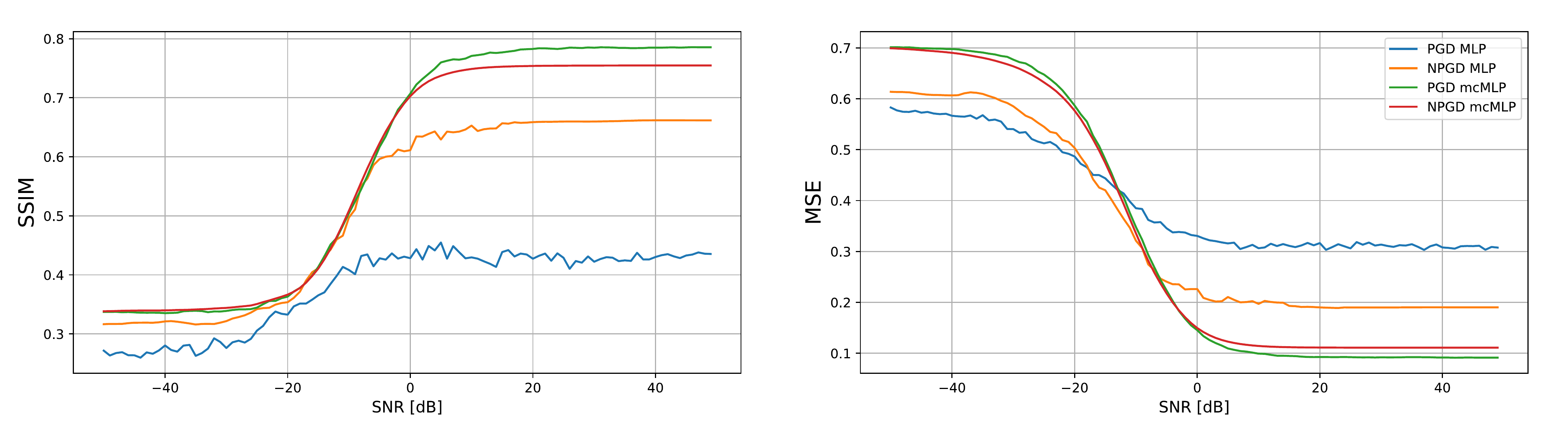}
\caption{\textbf{MNIST Reconstruction with MLPGAN using NPGD and PGD.} With subsampling ratio of 5\%. As a comparison to the DCGAN architecture, the reconstruction error of the MLPGAN architecture is much worse, showing that a DCGAN is more capable at representing the structure of the dataset since it considers only local dependencies of the pixels in the image.}\label{fig:mlp_losses_m39}
\end{center}
\end{figure*}

\onecolumngrid

\begin{figure}[H]
    \centering
    Subsampling Ratio 2\%\\
    \includegraphics[width=1\textwidth]{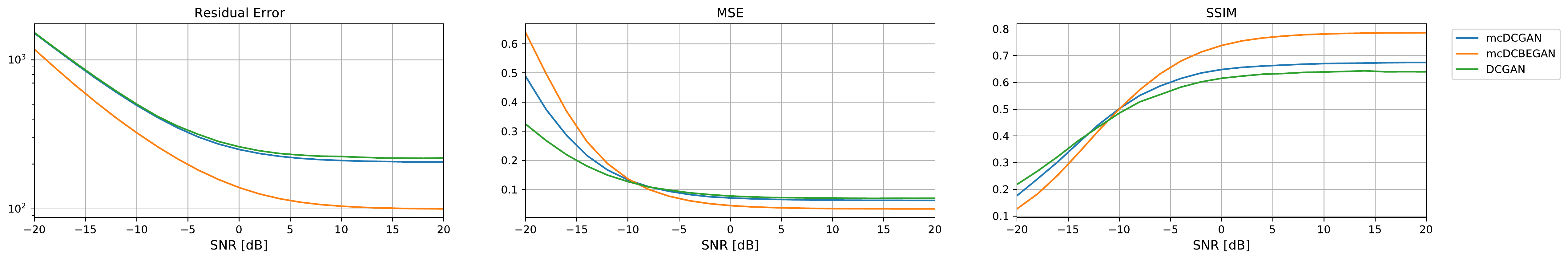}\\
    Subsampling Ratio 5\%\\
    \includegraphics[width=1\textwidth]{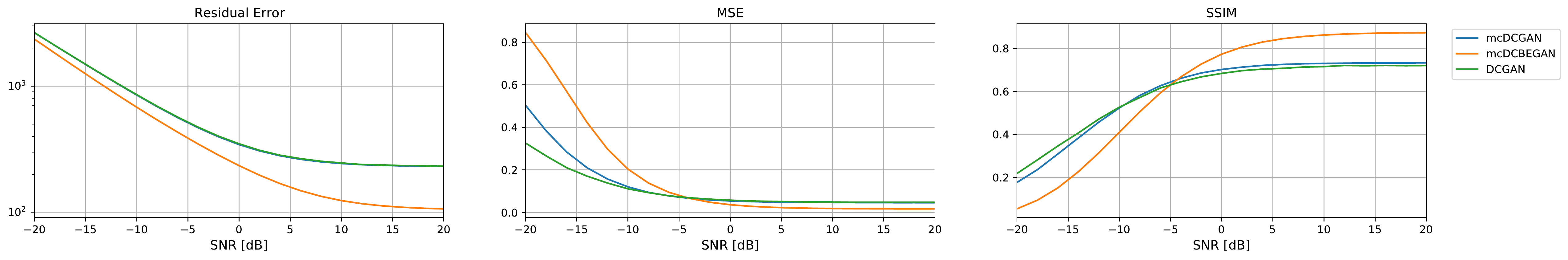}\\
    Subsampling Ratio 10\%\\
    \includegraphics[width=1\textwidth]{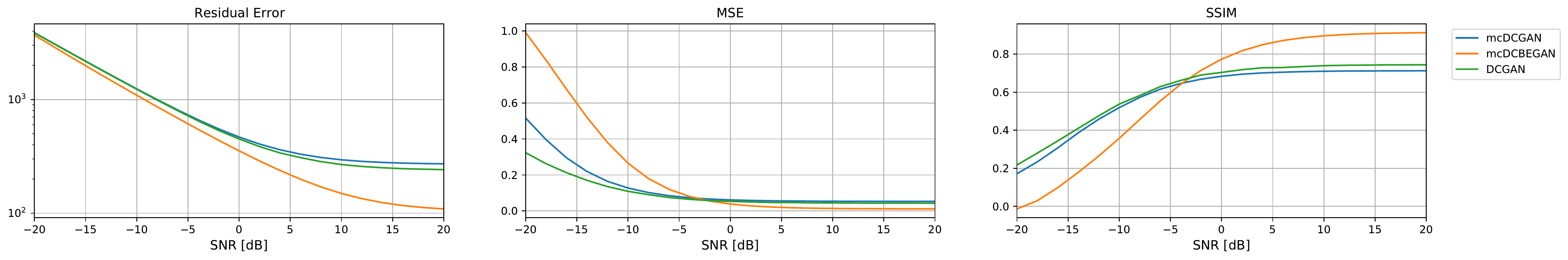}\\
    \caption{\textbf{CelebA Reconstruction Residuals using PDG} We trained the models for only 5 epochs because we had only a limited number of resources. Comparing the DCGAN and the mcDCGAN the conditioning on the measurement is not really improving our result. For small subsampling ratios the conditioning has still a small advantage due to having a much larger impact when there is few information, but for larger ratios the unconditional models are performing better. The missing performance difference which was expected between the unconditional and measurement conditional model is probably due to the relatively short training period and the training on the noiseless conditioning (\ref{eq:minMax}). For more complex data-sets the measurement condition can not just easily be induce and produce much superior performance. The limited training time could skew the results toward the unconditional models since it has much less parameters. The conditional GAN training is much more complex. Due to the auto-encoding abilities of the mcDCBEGAN the SSIM is much better since it produces visually closer but blurrier images (Fig. \ref{fig:CelebA_PDG_IMAGES}).  }
    \label{fig:CelabA_PDG_LOSSES}
\end{figure}

\begin{figure}[H]
\begin{center}
\includegraphics[width=1.\textwidth]{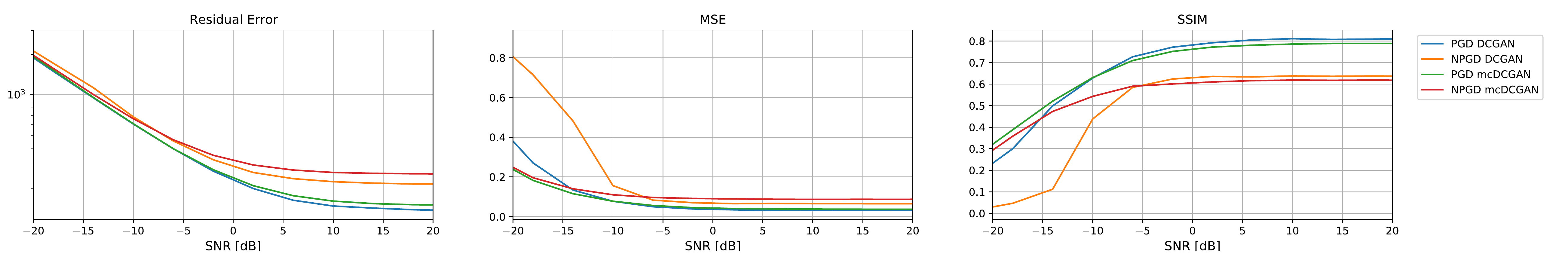}
\caption{\textbf{CelebA Reconstruction Residuals using PGD and NPGD.} The models were trained for 40 epochs. The subsamling ratio is 5\%. The NPGD is greatly reducing the performance for the unconditional models and small SNR, but the reconstruction is much faster with a speedup of $140-175$x in our experience. The performance difference between PGD and NPGD is due to projection error which can be bounded for a special case described in Theorem 2. The performance difference between PGD and NPGD is much closer for the measurement conditional models due to an reduced projection error using an conditional generator. }
\label{fig:celeba_losses_m614}
\end{center}
\end{figure}

\begin{figure}[t!]
    \centering
    \includegraphics[width=1\textwidth]{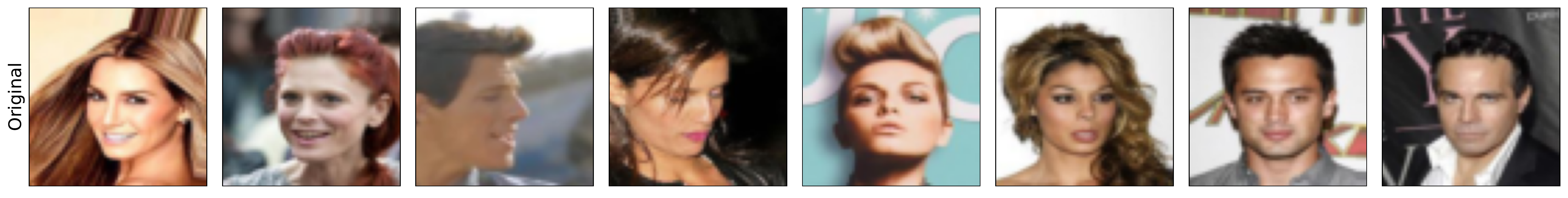}\\
    Subsampling Ratio ~2\%\\
    \includegraphics[width=1\textwidth]{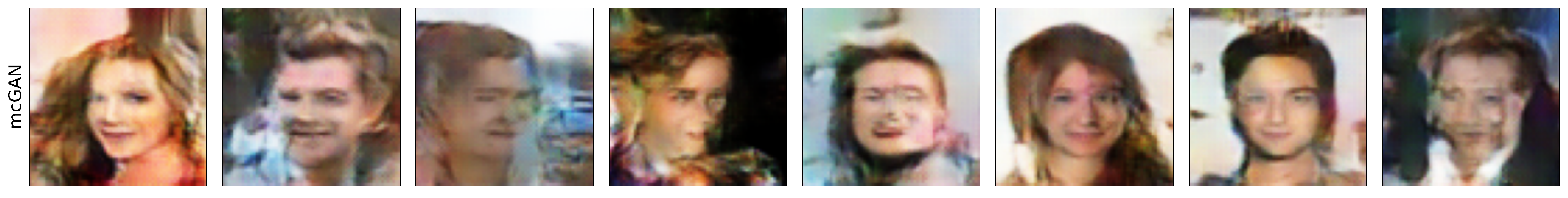}\\
    \includegraphics[width=1\textwidth]{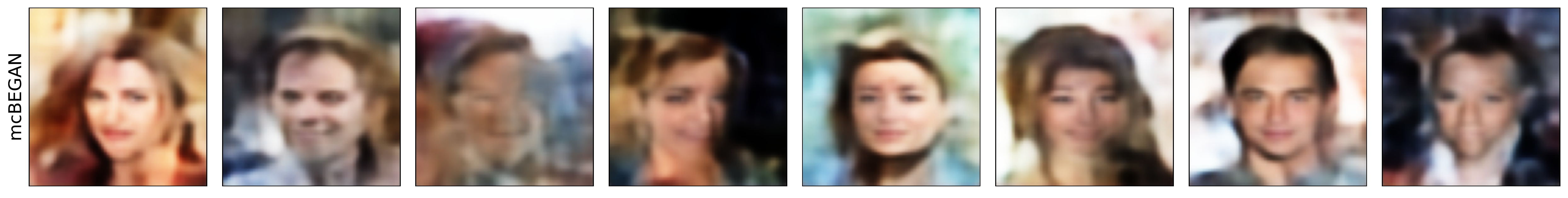}\\
    Subsampling Ratio ~5\%\\
    \includegraphics[width=1\textwidth]{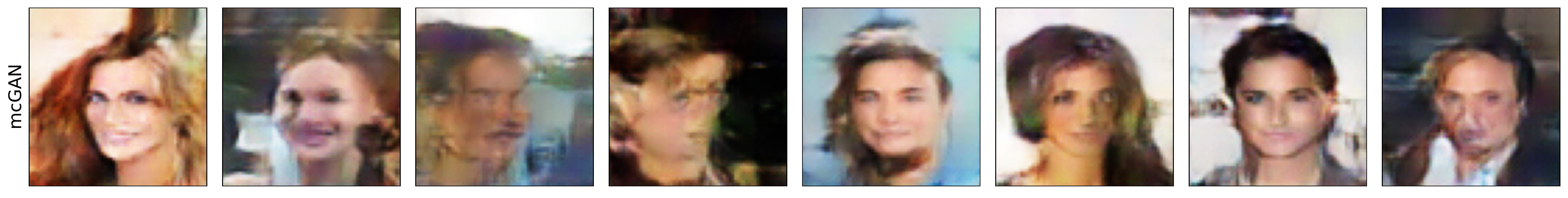}\\
    \includegraphics[width=1\textwidth]{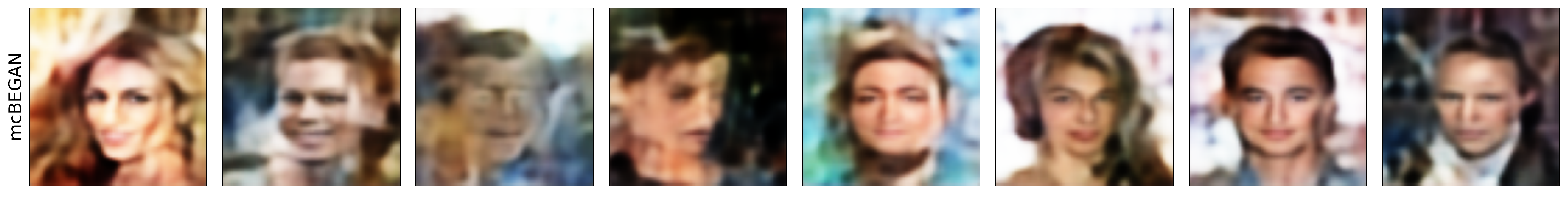}\\
    Subsampling Ratio ~10\%\\
    \includegraphics[width=1\textwidth]{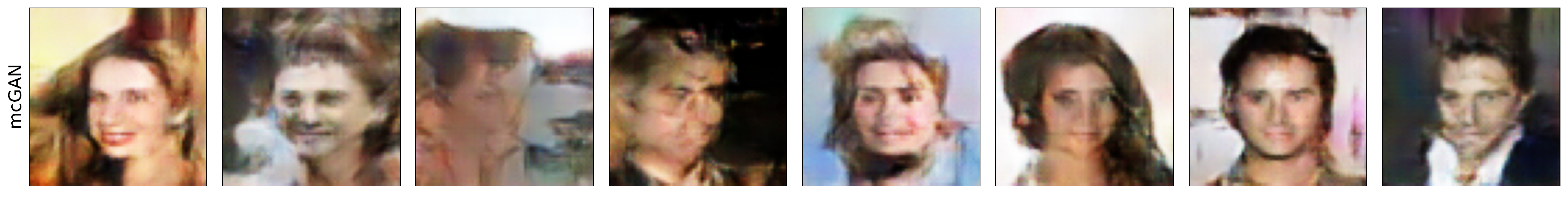}\\
    \includegraphics[width=1\textwidth]{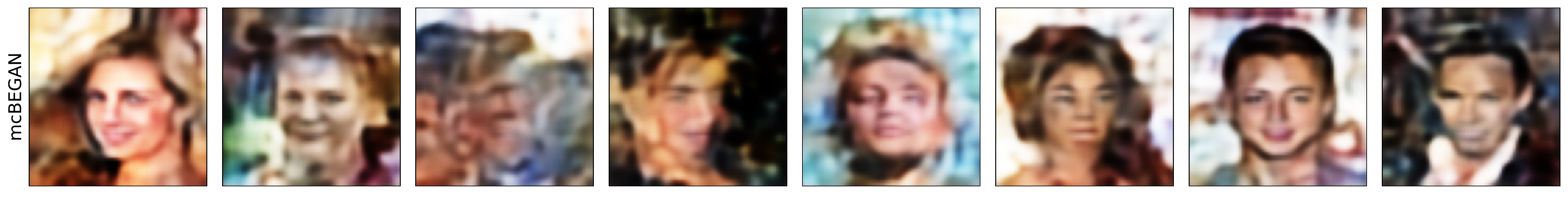}\\
    \caption{\textbf{CelebA Reconstruction Images with DCGAN using PDG.} In this figure the models are of the DCGAN architectures trained for 5 epochs. Comparing the mcGAN and mcBEGAN for a subsampling rate of 2\% 5\% and 10\% with a SNR of 0dB, the mcBEGAN is much closer to the original image. Due to its autoencoding capabilities, the mcBEGAN is able to produce blurred images that are still close to the original. The BEGAN balancing parameter was chosen to be $\gamma = 0.5$. Comparing the MSE 0dB (Fig. \ref{fig:CelabA_PDG_LOSSES}) all models have about the same performance with a small advantage for mcBEGAN. The SSIM is shows clearly an advantage for the mcBEGAN which is inline with the visual results.}
    \label{fig:CelebA_PDG_IMAGES}
\end{figure}

\twocolumngrid
\null\newpage
\null\clearpage

\section{Conclusion}
In this work, we combined two frameworks into a new reconstruction
approach for inverse networks. A measurement-conditional generative
model has shown to be improving the reconstruction accuracy in terms
of MSE loss for the MNIST dataset and the usage of pseudoinverse
networks (NPGD) has yielded in general a large reduction in
computation time, in our experience by a factor of $140\dots175$. At
the same time, the combination of the conditional models with the NPGD
algorithm still achieves almost as good results as PGD and for the
MNIST dataset even better results compared to the unconditional
models. The proposed method therefore allows for reconstructions in a much larger scale, but the advantage of using NPGD and measurement conditional models uses its full potential only in the reconstruction of many observed signals with the same measurement matrix, due to having a high upfront cost. 
The performance difference between the mcDCBEGAN and the mcDCGAN is small, whereby the mcDCBEGAN is slightly outperforming the mcDCGAN most of the time. 

Future work directions could be to apply additive noise to the measurements for training the conditional GANs, which stabilizes the training and possibly achieve improvements in reconstruction accuracy. Another possible direction for improving the accuracy is to investigate conditional pseudo-inverse networks, or combining the NPGD with the PGD algorithm, where initially using NPGD would result in an already close result and subsequently further improved by the introduction of PGD.

\section*{Acknowledgments}
We thank Jürgen Bauer for helpful discussion and comments.
\bibliography{quellen.bib}

\null\clearpage

\newpage
\onecolumngrid
\section{Appendix}
\begin{figure*}[h!]
\begin{center}
\includegraphics[width=1\textwidth]{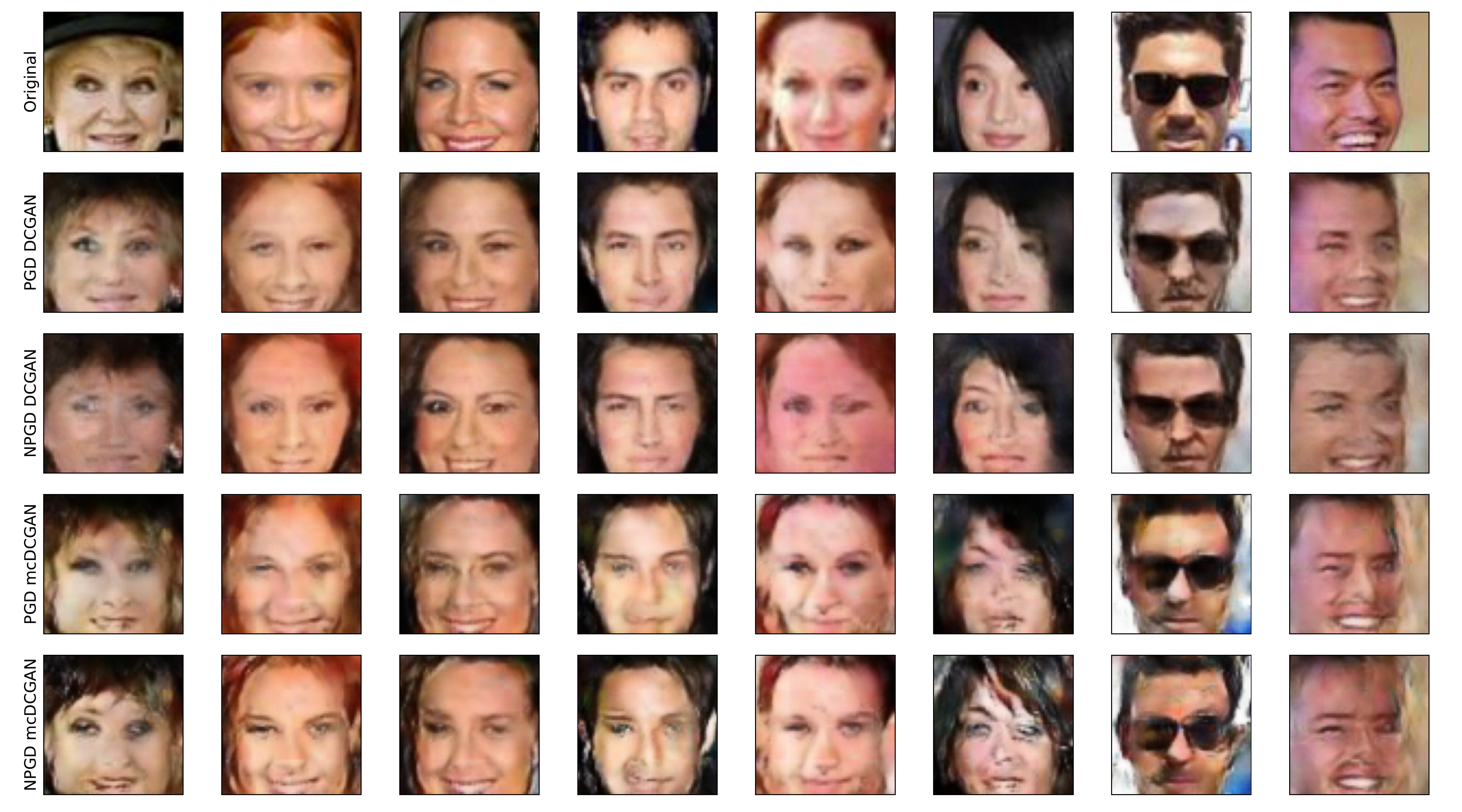}
\caption{\textbf{CelebA Reconstruction Images with DCGAN using NPGD and PGD.} Using a subsampling ratio of 5\% in the noiseless case. The DCGAN and mcDCGAN models are trained for 40 epochs. The losses in Fig. \ref{fig:celeba_losses_m614} are similar to the models trained for 5 epochs, hence the quality of the samples seems to be similar to the ones shown in Fig. \ref{fig:CelebA_PDG_IMAGES}. Moreover, these results for the noiseless case are comparable to the results achieved at the start of the convergence on the MSE which happens at around -5dB SNR.}
\end{center}\label{fig:celeba_images_m614}
\end{figure*}

\end{document}